\documentclass[11pt]{article}

\usepackage[margin=1in]{geometry}  
\usepackage{setspace}  
\usepackage{caption}
\usepackage{gensymb}

\usepackage[fleqn]{amsmath}  
\usepackage{amssymb}
\usepackage{empheq} 
\usepackage{bm,upgreek}  

\usepackage{changepage}

\usepackage{url}

\usepackage{algorithmic,algorithm}  

\usepackage{graphicx}  
\usepackage{xspace}

\usepackage{color}  

\usepackage{tikz}
\usetikzlibrary{fit,positioning}
\usetikzlibrary{bayesnet}
\usepackage{wrapfig}

\usepackage{footmisc}
\newcounter{secnum}
\usepackage[all]{nowidow}

\usepackage{titlesec}
\titleformat*{\section}{\Large\bfseries}
\titleformat*{\subsection}{\onehalfspacing\Large}
\titleformat*{\subsubsection}{\large\bfseries}
\titleformat*{\paragraph}{\large\bfseries}
\titleformat*{\subparagraph}{\large\bfseries}

\usepackage{fancyhdr}
\usepackage{indentfirst}
\usepackage{lipsum}  

\usepackage{adjustbox}
\let\oldhash\#%
\DeclareRobustCommand{\#}{\adjustbox{valign=B,totalheight=.35\baselineskip}{\oldhash}}%

\newfont{\namefont}{cmr10 at 12.5pt}

\begin{document}
\vspace*{0cm}
\begin{center}
\noindent{\LARGE Frontiers in Evolutionary Computation:\\[.25cm]  A Workshop Report}
\vspace{.5cm}

\rule{0.75\textwidth}{.4pt}
\vspace{.5cm}

\begin{minipage}{.3\textwidth}
\centering
\onehalfspacing
{\namefont Tyler Millhouse}\\ Santa Fe Institute\\ tyler.millhouse@santafe.edu\\
\end{minipage}
\begin{minipage}{.3\textwidth}
\centering
\onehalfspacing
{\namefont Melanie Moses}\\ University of New Mexico\\ melaniem@cs.unm.edu\\
\end{minipage}
\begin{minipage}{.3\textwidth}
\centering
\onehalfspacing
{\namefont Melanie Mitchell}\\ Santa Fe Institute\\ mm@santafe.edu\\
\end{minipage}\vspace{.5cm}

\rule{0.75\textwidth}{.4pt}
\vspace{.5cm}
\begin{quote}
\textbf{Abstract:} In July of 2021, the Santa Fe Institute hosted a workshop on evolutionary computation as part of its Foundations of Intelligence in Natural and Artificial Systems project. This project seeks to advance the field of artificial intelligence by promoting interdisciplinary research on the nature of intelligence. The workshop brought together computer scientists and biologists to share their insights about the nature of evolution and the future of evolutionary computation. In this report, we summarize each of the talks and the subsequent discussions. We also draw out a number of key themes and identify important frontiers for future research.
\end{quote}

\end{center}

\newpage
\tableofcontents
\newpage
\section{Overview}

In developing a deeper understanding of intelligence, it would be a serious mistake to ignore the only design process which has lead to general intelligence: biological evolution. Upon examining this design process, one is struck by the diversity and competence of evolved organisms and by the remarkable simplicity and short-sightedness of evolution itself. These observations suggest a promising role for evolutionary computation in developing artificial general intelligence. This approach is so promising because it does not require human investigators (or anything else) to have a complete picture of how to build agents with general intelligence. That said, we must be cautious when emphasizing the simplicity of evolution as a design process. Evolutionary mechanisms like natural selection \textit{are} strikingly simple when compared to the complex systems evolution has built, but this observation neglects the rich biological and ecological contexts in which evolution acts.

If this were not the case, then the job of developing and applying evolutionary algorithms should be fairly straightforward. However, in a recent paper, Risto Miikkulainen and Stephanie Forrest (2021) outline a number of areas in which evolutionary computation lags behind biological evolution. The advantages of biological evolution include its open-endedness, its capacity for major design transitions, its ability to exploit neutral evolution and co-evolution, its implicit multi-objectivity, and its complex genotype to phenotype mappings. While these are natural targets for research in evolutionary computation, we do not sufficiently understand which features of biological evolution are responsible for these advantages or how to incorporate these features into useful evolutionary algorithms. The aim of this workshop was to bring together researchers in biology and computer science to gain a better understanding of both biological and simulated evolution and to identify new frontiers for research in evolutionary computation.

\section{Summaries of Talks and Discussions}

\subsection[``Does Computation Make Us Blind?'' (Rodney Brooks)]{``Does Computation Make Us Blind to Essentials of Biological Evolution?''}

\begin{adjustwidth}{1cm}{1cm}
\onehalfspacing
\textit{Rod Brooks is the Co-Founder and Chief Technical Officer of Robust.AI. He previously served as the Panasonic Professor of Robotics at MIT and the Director of the MIT Computer Science and Artificial Intelligence Laboratory. His work addresses subjects such as robotics, artificial intelligence, and artificial life.}
\end{adjustwidth}


\noindent In the opening talk of the workshop, Rod Brooks questioned the metaphors we commonly use to understand intelligence. In particular, Brooks questions the value of \textit{computation} as a metaphor for understanding the activity of intelligent systems. These doubts would not be controversial in other fields. For example, engineers and physicists working on new rocket engines do not find it useful to characterize the activity of a rocket engine as computation. The same cannot be said for cognitive and computer scientists seeking to understand intelligence. In this domain, it seems, a great deal of progress has been made by the use of the computation metaphor. Why then should we consider replacing it?

Brooks begins by noting that our concept of computation is largely an outgrowth of our own cognitive abilities (e.g., the ability to follow rules and manipulate symbols). Indeed, the term ``computer'' first referred to \textit{people} who performed computations by hand, and algorithms (e.g., for finding square roots) were developed long before modern computers. Even where computers exceed our computational abilities (e.g., in memory), engineers have been limited by what they can physically build. For these reasons, Brooks argues, we should think of our concept of computation as largely culturally constructed. Hence, we cannot assume that computation is a primitive element of the world---much less the best way to understand intelligence. Instead, we should pragmatically consider which conceptual framework is best for advancing this understanding. 

Fortunately, computer and cognitive science are not alone in their need for new conceptual frameworks or, as Brooks calls them, ``Newtonian black boxes." Conceptual innovation is an important part of how sciences mature (see e.g., the development of calculus or the replacement of phlogiston theory in chemistry). A feature and a bug of any Newtonian black box is that it can be applied to a wide range of circumstances where the minimal conditions for applying it are met. This allows us to generalize our understanding and ignore irrelevant details, but it also raises the possibility that we have over-extended our concept and are missing vital details. 

Hence, even if brains seem like apt targets for a computational description, we should be vigilant about missing details and be willing to adopt new black boxes whenever necessary. Brooks appeals to cellular automata to illustrate how a Newtonian black box can mislead us. In particular, he considers a cellular automaton which gives rise to interesting spatial structure when \textit{and only when} the space of cells is toroidal (i.e., when we connect the edges to each other, ensuring that edge cells have neighbors on all sides). The exact nature of this structure is not important. What matters is that if we focus only on analysing the transition rule of the automaton, we will not be able to understand or anticipate the structure that emerges. 

In summary, Brooks argues that metaphors we use when studying intelligence may play a decisive role in whether we will ultimately understand the nature of intelligence. Computation is not, he argues, a primitive element of the world, but a concept shaped by our own experiences and limitations. As such, we should reconsider whether computation is the right metaphor for understanding intelligence or whether new metaphors are needed.

\noindent \textbf{Discussion:}

The group discussion began with a request for Brooks to clarify the sense in which he thinks computation is socially constructed. Brooks clarified that it is not so much computation in some abstract sense that is constructed, but rather our favored models of computation. Turing machines, for example, are but one formalism for thinking about computation. Humans have found this an intuitive way to reason about computation, but that does not mean that it is the model of computation that will best illuminate the nature of intelligence. The discussion next turned to Brooks's thoughts on evolutionary computation. Brooks noted that his previous statements on the matter (e.g., that evolutionary computation has a 50\% chance of growing to dominate the field of AI) were not meant to single evolutionary computation out as his favored approach, but rather to say that it or some other revolutionary approach will be required to substantially advance our understanding of intelligence. Several commenters also noted examples of the kind of transition of metaphors recommended by Brooks. For example, fully connected neural networks and convolutional neural networks can compute the same functions, but in practice the latter seem much better suited to a number of tasks (e.g., image recognition). Hence, we might be able to think of this transition in terms of a switch from one model of neural computation to another. Another example might be the earlier transition from mathematical to computational models in science. Several commenters also asked what Brooks thought the next paradigm in AI might be. Brooks left the issue open, but one commenter suggested that our existing models of computation are optimized for doing arithmetic and that a new model optimized for agents and the relations between them might be more illuminating.

\subsection[``The Evolutionary Paradox of Robustness'' (Steve Frank)]{``The Evolutionary Paradox of Robustness, Genome Overwiring, and Analogies with Deep Learning''}

\begin{adjustwidth}{1cm}{1cm}
\onehalfspacing
\textit{Steve Frank is Donald Bren Professor of Evolutionary Biology and UCI Distinguished Professor at the University of California, Irvine. His current research concerns the history of microbial life, the evolution of regulatory control, and invariance in common natural patterns.}

\end{adjustwidth}
Steve Frank described the \textit{paradox of robustness} and argued that it might help us to understand distinctive features of genomes and deep neural networks. The paradox of robustness describes a situation where increasing the robustness of a mechanism results in a reduction in robustness for its components. For example, a redundant array of inexpensive disks (or ``RAID'') pairs a RAID controller with several hard disks. Whereas a single hard disk is susceptible to failure and permanent data loss, the RAID controller stores data redundantly across multiple disks. If a single disk fails, it is replaced, and the data is restored from the remaining disks. The result is that individual disk failures carry a much smaller risk of data loss. This robustness to loss allows users to economize on the cost and quality of the individual disks. Hence, the ``inexpensive'' in redundant array of inexpensive disks. 

This is paradoxical since increased robustness at the global level is accompanied by a reduction in the robustness of individual parts. There are plausible biological analogs of this phenomenon. For example, cold-blooded animals must use proteins that perform their function reliably across a wide range of temperatures, since their body temperatures fluctuate significantly with changes in their environment. The same is not true for warm-blooded animals whose higher-level temperature control system keeps their body temperature within a much smaller range. As a result, temperature-robust proteins will confer little or no fitness advantage in warm-blooded animals. Hence, if a less robust variant is substantially cheaper to produce, selection will favor that variant despite the loss of robustness. As in the hard disk case, a higher-level control system that increases robustness results in a loss of robustness for the individual parts. 

With this foundation in place, Frank argues that the paradox of robustness might explain  puzzling phenomena in both biology and machine learning. A key element of the paradox is the transition from reliance on robust lower-level parts to robust higher-level control. In biology, networks of genes perform regulatory functions in organisms (e.g., the regulation of protein production, metabolism etc.). These networks, according to Frank, appear to be somewhat overbuilt with more genes and more links between them than one might expect given their function. Perhaps we could explain this observation by proposing that evolution increasingly relied on higher-level regulatory networks in order to economize on the reliability the of systems they regulate. 

Similarly, deep neural networks often perform very well while being over-parameterized. Curiously, the performance of these networks on unseen data often continues to improve even after their performance has leveled off on their training data. One might explain this phenomenon by proposing that as training continues, the network shifts from memorizing/overfitting the data (which is costly in terms of model complexity and punished by regularization) to relying on higher-level representations by which the network can organize and interpret lower-level representations. This does little to improve performance on training data, but it does mean that the network has learned a more generalizable solution and so performance on unseen data improves. 

While these proposals are admittedly speculative, Frank suggests that this phenomenon might be endemic to design processes like evolution or gradient descent. Understanding this phenomenon can help us to explain curious design choices in nature and better exploit optimization processes in machine learning. For example, another aspect of the paradox of robustness is its irreversibility. Once the components have become less robust, it is difficult or impossible for the system to discard its high-level control system and rely again on the robustness of the individual components. Noting this fact could allow us to anticipate and (where necessary) head off the permanent loss of functionality in systems designed by evolutionary algorithms. 

\noindent\textbf{Discussion:} 

An important topic of discussion was where the paradox of robustness might apply to biology and how it might manifest differently in evolutionary computation. One questioner noted that in simulated evolution, solutions tend to be persistent once they are learned despite continued learning and improvement. Frank suggested that costs in biological evolution are more pervasive than in evolutionary computation. For example, there is a cost to maintaining each gene or producing each protein, and this would imply a more systematic pruning effect in biological evolution. Returning to biology, Frank suggested that DNA might be less strongly pruned than proteins---creating a situation where new regulatory genes will be favored if they can reduce the need to produce costly proteins. As one commenter proposed, the genetic regulatory network is the control system which takes on an increasingly vital role as evolution seeks to economize on proteins at the expense of more regulatory genes. 

Another important subject of discussion was the balance between robustness and variation in replication. As Frank suggests, more robust replication is often favored in the short term where preserving a useful function is key. More variation, however, is often favored in the long term where a species must adapt to environmental changes. To further complicate matters, the risks and rewards of variation may differ for different genes. As one commenter suggested, striking the right balance between variation and robustness of replication might impact which regulatory systems are most adaptive. For example, when subject to genetic mutations, some might be less susceptible to catastrophic failures or be more likely to produce potentially adaptive changes in function. Along similar lines, another commenter suggested (on the basis of research on artificial neural networks) that having systems with considerable redundancy in their constituent parts might result in a fitness landscape more conducive to adaptive evolution. Frank echoed these thoughts, arguing that a better understanding of the geometry of fitness landscapes could lead to major advances in our understanding of evolution.

\subsection[``Open-Endedness, Quality Diversity, and Deep Learning'' (Ken Stanley)]{``Open-Endedness, Quality Diversity, Deep Learning, and the Future of Evolutionary Computation''}

\begin{adjustwidth}{1cm}{1cm}
\onehalfspacing
\textit{Ken Stanley is the Open-Endedness Team Leader at OpenAI. He previously served as Charles Millican Professor of Computer Science at the University of Central Florida. Ken's work focuses on subjects such as open-endedness in evolution and the evolution of neural networks.}
\end{adjustwidth}

Ken Stanley explored the idea of \textit{open-endedness} as a feature of evolution and considered diverse ways of incorporating open-endedness into AI programs. Evolution is clearly a powerful optimization strategy, and it is responsible for the complexity and diversity of living organisms. Nevertheless, evolutionary algorithms often take a back seat to gradient descent, with many machine learning domains dominated by large neural network models. These observations prompt two questions: What is responsible for the remarkable success of evolution in nature, and how can we transfer its virtues to other optimization algorithms (e.g., evolutionary algorithms)? 

Stanley argues that a key advantage of natural evolution is its open-endedness, and that this feature is largely absent in existing machine learning algorithms. Open-endedness is the tendency of a design process to continually increase the complexity and diversity of the systems upon which it acts. As Darwin famously observed, the ongoing process of evolution produces``endless forms most beautiful'' (Darwin, 1872, p. 429). This feature of evolution is not only aesthetically satisfying, it contributes heavily to the surprising, diverse, and often ingenious solutions discovered by evolution. For this reason, Stanley seeks to understand what is responsible for the open-endedness of natural evolution and how we might make other machine learning algorithms more open-ended. One approach would be to encourage algorithms to produce novel solutions (e.g., by using an objective function that rewards novelty). This might work for some applications, but it is doubtful that this is the secret to the success of natural evolution, which does not directly detect and reward novelty. As one commenter noted, it seems optimal to focus on ways of getting open-endedness and novelty to emerge naturally. 

Stanley next considers several less direct approaches. First, the environment seems to play an important role in the generation of novelty. For example, agents are part of their own environment, and the opportunities available to an agent in a given environment are shaped by the nature of the agent itself. As the agent evolves, new opportunities naturally arise. Environments can also evolve and change on their own. This is especially true when agents are co-evolving with partners and adversaries. To model this phenomenon, Stanley devised a simulation in which mazes co-evolved with solutions to mazes. The result was a population of solutions and mazes that consistently increased in complexity and diversity.

Second, diversity can be promoted by avoiding aggressive optimization and by determining fitness through the satisfaction of some minimal criterion. These ``good enough'' solutions more readily proliferate and diversify, whereas rewarding the ``best'' solution can lead to premature convergence. For example, in the maze/solution co-evolution program, solutions were allowed to reproduce so long as they solved at least one maze, and mazes were allowed to reproduce as long as they were solved by one solution.

Third, Stanley argues that resource limitation can play an important role in promoting open-endedness. Consider the difference between an environment with one particularly abundant food source and an environment with several significantly less abundant sources. For a foraging species in the first environment, there is a greater chance that the entire population will specialize in foraging for a single food source. In the second environment, the ability to exploit multiple food sources would be favored. Analogously, in the maze/solution co-evolution program, Stanley capped the number of solutions that could reproduce by solving any particular maze. This meant that the solutions could not converge on solving the easiest maze and the mazes could not converge on being the easiest to solve. This led to more diverse and difficult mazes being evolved as solutions explored other mazes. 

It is important to note that (i) agent-environment co-evolution, (ii) minimal criteria, and (iii) resource limitation can be implemented in evolutionary algorithms and non-evolutionary algorithms. These features, thus, have the potential to bridge the gap between different kinds of algorithms. As Stanley argues, it is not so important for our final algorithm to be unambiguously evolutionary if we can transfer the principal advantages of evolutionary algorithms to a wider range of approaches.  

\noindent\textbf{Discussion:} 

The first issue raised in discussion was the concern that having to employ a mechanism that explicitly enforces diversity indicates some failure in the approach. Diversity, the commenter argued, should emerge spontaneously from well-designed evolutionary algorithms. Stanley essentially agreed with this concern, but noted that mechanisms like resource limitation and minimal criteria are intended to help diversity emerge without it being explicitly rewarded. Another commenter worried that these kinds of mechanisms might not result in long-lasting diversity, but might in the long run lead to convergence or to no new diversity being generated. Stanley argued that diversity for diversity's sake may not be what we ultimately want in any case. What we want is a process to continue producing \textit{interesting} solutions to problems. We can design algorithms with this in mind, but, as yet, diversity has been easier to formalize and, hence, has been a reasonable proxy for interestingness. That said, Stanley also argued that a major limitation of existing algorithms is imposed by the kinds of environments we are able to simulate. For example, mazes might increase in diversity, but they will remain fundamentally a maze. Natural environments are far richer, and trying to bring some of this richness to evolutionary computation could allow for the continued production of diversity. 

Other commenters raised questions about what we should focus on when thinking about evolutionary computation. One asked to what extent we could explain the emergence of diversity by focusing on how changes to individuals affect the kinds of strategies they can employ and create opportunities for further change and diversification. Stanley agreed that this was an important point, but stressed the relationship of these individuals with other agents and with their own environment. A second commenter asked to what extent the algorithms Stanley described blurred the lines between ``evolution'' in the colloquial sense of incremental improvement and ``evolution'' in the biological sense. Stanley argued that he prefers to think of evolution in the biological sense as one of a larger family of open-ended learning algorithms. For this reason, he thinks we can study the mechanisms that underlie open-endedness without clearly distinguishing evolutionary algorithms from other approaches. 

\subsection[Discussant Comments (Dave Ackley)]{Discussant Comments \& General Discussion:}

\begin{adjustwidth}{1cm}{1cm}
\onehalfspacing
\textit{Dave Ackley is an emeritus professor of Computer Science at the University of New Mexico. Dave's research involves neural networks, evolutionary algorithms, artificial life, and biological approaches to computing.}
\end{adjustwidth}

In his commentary on the first day's talks, Dave Ackley argued that embodiment was central to the issues raised in each of the talks. More generally, Ackley argues that thinking about embodiment is the answer to many (if not all) of the challenges that face evolutionary computation. The importance of embodiment was explicit in Rodney Brook's talk, but the other talks (argued Ackley) centrally concern embodiment as well. For example, Steve Frank argued that the biological costs of regulatory mechanisms and proteins might lead to distinctive design choices in biological organisms. Ken Stanley used resource limitation and minimal criteria to improve diversity, but both of these strategies involve a more realistic representation of the constraints and selective mechanisms faced by organisms in real physical environments. 

The general discussion that followed Ackley's remarks raised several important issues. One commenter questioned Ackley's emphasis on embodiment over the more general notion of situatedness, noting that work on adversarial learning for computer network security does not consider embodied agents but does consider the fact that agents are situated in a dynamic adversarial context. Ultimately, Ackley argued that while one could think in these terms, the notion of situatedness seemed to emphasized factors outside the organism while embodiment focused on factors internal to the organism.  

Another commenter worried that Ackley's optimism about embodiment might be misplaced if the main limitation of evolutionary algorithms is that they can't run for evolutionary time (e.g., billions of years). In this case, it is not clear we should expect embodiment (or any other design variation) to fundamentally advance evolutionary computation since the evolution of animal bodies built upon innovations developed over billions of years of microbial evolution. Ackley acknowledged the general point but argued that we may nevertheless be able to make progress on more modest applications of evolutionary algorithms. To do so, he argues that we must be clear about the shortcuts taken and how they impact the scalability and environmental appropriateness of our algorithm.

\subsection[``Embodying the Products of Evolutionary Computation'' (Josh Bongard)]{``Embodying the Products of Evolutionary Computation''}

\begin{adjustwidth}{1cm}{1cm}
\onehalfspacing
\textit{Josh Bongard is the Veinott Professor of Computer Science at the University of Vermont and Director of the Morphology, Evolution \& Cognition Laboratory. His research involves the design, manufacture, and evolution of robots and biological organisms.}
\end{adjustwidth}

Josh Bongard detailed several frontiers for artificial intelligence that involve evolutionary processes. Bongard argues that a key feature of biological evolution is its facility for ``acting on and making use of the richness of spontaneous physical order.'' The idea here is that under appropriate circumstances, physical systems self-organize in interesting and potentially useful ways. For example, the branching pattern of flowing water in a river delta arises spontaneously from the natural behavior of water as it follows and shapes the gradient of the surrounding land. A similar branching pattern is formed spontaneously when the direction of root growth tracks moisture gradients in the surrounding soil. In both cases, no planning is required to ensure that water finds its way to the sea or that roots find water in the soil. 

Bongard argues that finding better ways to understand and exploit spontaneous physical order is an important frontier of AI research. More broadly, Bongard contends that we should start with spontaneous physical order, see what kinds of problems it can solve, and only introduce computation where needed. Here Bongard echoes Brooks's skepticism about the centrality of computation in AI. This project might take the form (following nature) of evolving solutions that exploit order (e.g., by evolving virtual organisms in physically realistic contexts where natural order emerges and can be exploited). It might also take the form of exploiting solutions already developed by biological evolution. Bongard's work involves both of these approaches. For example, in his work on ``xenobots,'' clusters of animal cells (e.g., skin or muscle cells) can be joined into biological robots which exhibit surprising and self-organizing behaviors. These xenobots can also be simulated, and novel arrangements of different cell types can be evolved in physically realistic virtual environments before being constructed using actual living cells. Once constructed, they often exhibit the same adaptive behaviors selected for in the simulation. In this work, Bongard not only evolves novel arrangements of cells but does so in a way that relies on the cells' previously evolved machinery. 

Bongard also identifies three other important frontiers for work in this area. First, the process of transforming designs evolved \textit{in silico} to real world xenobots is often laborious, and not all evolved designs can be physically realized. Further research on improving and automating this process would allow for faster and less constrained design iteration. Another important frontier is finding the best roles for evolutionary and gradient-based optimization when designing solutions \textit{in silico}. Gradient-based methods have many advantages, but they rely on our ability to compute specific design tweaks from whatever our measure of success tells us about the current design. For example, how could we compute specific design tweaks to the body of a virtual fish from a measure of its forward speed while swimming? Only recently have researchers identified ways of doing this and even then only when design choices are artificially limited. This leaves open an important role for evolutionary design processes which can help us to explore a more diverse space of possible solutions. 

The last frontier concerns the use of evolution to create more \textit{transparent} xenobots. One difficulty for many methods in AI and machine learning is ensuring that the solutions they discover are comprehensible to human researchers. In the case of xenobots, Bongard suggests that we evolve experimental interventions, selecting those for which competing hypotheses suggest the most easily distinguishable predictions. For example, at what angle should a cut in a xenobot be made in order to test competing hypotheses about its repair processes? Using a simulation of these processes, different angles can be tried under different hypothesized repair processes. Those which yield the largest differences in behavior can be selected and refined. 

All these proposed frontiers highlight an important theme of Bongard's talk and the workshop as a whole. We can think of evolutionary processes as searching a space of possible solutions which is constrained in various ways (e.g., physical laws, material properties, available developmental mechanisms, etc.). By ignoring these constraints, purely computational approaches  have the apparent advantage of exploring a wider and more diverse space of possible solutions. Nevertheless, it seems clear that searching this space will be much more effective if we can somehow limit the possible solutions to those which tend to be more promising. A key difficulty, then, is finding ways of helping search along without unduly restricting the space of possible solutions. Bongard's recommendation to first exploit naturally-occurring order and only resort to computation where necessary is one strategy for restricting our search to more promising solutions. Of course, fully adopting this approach will require a better understanding of the physical and developmental processes at play in biology and how they further evolution by creating opportunities for evolution to take advantage of naturally-occurring order.

\noindent\textbf{Discussion:} 

The discussion focused on how to understand and improve the process of simulating evolution and then physically realizing the result. For example, one commenter asked if the evolution simulation could be constrained to only evolve organisms that could be physically constructed with existing technology. Bongard acknowledged the issue as one they have worked on and continue to work on, noting that early work resulted in many designs that could not be realized. Bongard's collaborator, Michael Levin, added that the cells themselves might be induced to accomplish much of the necessary construction if their capacity for self-organization could be better exploited. 

Another commenter suggested that it might be helpful to think about exploiting natural order as exploiting \textit{invariances} in nature. For example, any shape rotated $360\degree$ will sweep out a circle. Such simple invariances might be exploited by organisms---especially during development. In a similar vein, one comment proposed evolving the plan of development instead of the adult phenotype. Bongard and Levin noted that understanding how the development process might be directed is an important and active area of research. One commenter raised the more fundamental question of when it becomes easier to drop the simulation and simply conduct evolutionary experiments in the real world. Bongard argued that real world experiments, exact simulations, and approximate simulations accelerated via deep neural networks would likely all be required to make significant progress.  

\noindent 

\subsection[``Evolutionary Biology and Random Drift'' (Mike Lynch)]{``Evolutionary Biology and Random Drift''}

\begin{adjustwidth}{1cm}{1cm}
\onehalfspacing
\textit{Mike Lynch is a professor at Arizona State University and Director of ASU's Biodesign Center for Mechanisms of Evolution. His work investigates the role of mutation, random genetic drift, and recombination in evolution.}
\end{adjustwidth}

\noindent Mike Lynch discussed the important role and surprising consequences of random drift in biological evolution. For a mutation to be favored by natural selection and (eventually) go to fixation in a population, that mutation must confer a fitness advantage on its possessors. It is easy to see that the size of this advantage is related to how quickly the mutation will go to fixation, but it is also related to \textit{whether} the mutation will go to fixation. In a population of infinite size, an arbitrarily tiny fraction of the population can possess a gene, and the death of any single individual cannot erase the gene from the population. In a population of finite size, the death of a single individual can (in principle) remove a gene from the population. More generally, whether a gene finds itself in such a precarious position will depend more on the fates of individual organisms when the population is small. The vicissitudes of life (and death) are such that individual survival is a chancy affair, one influenced but not determined by fitness. 

These random processes lead to changes in the frequency of alleles in a population in the absence of selective pressures. This \textit{random drift} can even swamp selective pressure in favor of a particular allele when that pressure is sufficiently small. All this means that in smaller populations where the effect of drift is greater, a gene must confer a greater fitness advantage in order for natural selection to drive that gene to fixation. This, incidentally, helps to explain why organisms with small population sizes (e.g., mammals) tend to have more genetic bloat than organisms with very large population sizes (e.g., bacteria). The cost of an extra bit of DNA is simply too small to be noticed by evolution. 

Population size and random drift also interact with genetic co-evolution and recombination. Recombination generally selects alleles from chromosomes in a random fashion---with alleles of the same gene on different chromosomes enjoying the same chance of being selected. That said, as sections of each chromosome are randomly selected and copied for inclusion in gametes, nearby alleles are unlikely to be separated. Such linkages allow for the possibility of genetic co-evolution and the weakness of a linkage reduces the effective population size for the relevant genes. For example, if one pair of alleles is better than another, that pair must persist long enough to be seen and favored by selection. Of course, the chance of separation by recombination is akin to the death of a mutation-possessing individual. As the degree of linkage can vary widely from pair to pair, linked pairs may even have to compensate for deleterious mutations that accumulate in unlinked pairs due to the latter's higher threshold for visibility to natural selection. 

While Lynch was hesitant to speculate about how these ideas might be applied to evolutionary computation, it seems clear that they suggest important principles for designing evolutionary algorithms. Parameters such as genetic linkage and population size are squarely within the control of researchers. Perhaps careful experimentation with these can yield more effective evolutionary algorithms. 

\noindent\textbf{Discussion:} 

Lynch's talk sparked a considerable amount of curiosity about how the ideas he presented might be applied to evolutionary computation. While Lynch largely deferred to computer scientists to work out possible applications, several suggestions arose during the discussion. For example, one commenter wondered whether idealized models in evolutionary biology might be tested and applied in evolutionary computation, where the factors that undermine the idealizing assumptions of the model are within the researcher's control. Another commenter asked about how the work might be applied to promote the evolution of diversity (hearkening back to Ken Stanley's talk). Lynch suggested that if diversity depends on co-evolution of different traits by the accumulation in individuals of groups of genes, then the recombination rate will affect the ability of these groups to persist long enough to be selected. Finally, another commenter asked about the extent to which evolutionary biologists might be able to optimize an evolutionary algorithm to maximize the speed of evolution. Lynch suggested that while this was possible in principle, it might require knowing (in some detail) the mapping from genotypes to phenotype. Yet, as the commenter pointed out, this mapping is far more accessible to those working in evolutionary computation since this mapping must be made explicit in order to be incorporated into an evolutionary algorithm.

\subsection[``Simulated Evolution on Static Functions'' (Darrell Whitley)]{``Simulated Evolution on Static Functions with Bounded Nonlinearity''}

\begin{adjustwidth}{1cm}{1cm}
\onehalfspacing
\textit{Darrell Whitley is Professor of Computer Science at Colorado State University and the Editor-in-Chief of the journal Evolutionary Computation. His research studies evolutionary computation, search optimization, and machine learning.}
\end{adjustwidth}

In his talk, Whitley focused on methods for improving the optimization abilities of evolutionary computation methods. This work was developed without considering any constraints from biological evolution, but some of the methods end up having possibly surprising connections to biology. Whitley focuses on a particular kind of optimization problem: K-bounded pseudo-Boolean functions. These are functions that map bit strings, $x$, to real values, and can be written as the sum of $M$ subfunctions, each of which depends only on $K$ bit positions in $x$. This class of functions subsumes many important real-world optimization problems. In fact, Whitley points out that this class of functions is NP-Complete, meaning that any class of problems in NP (e.g., satisfiability problems or Traveling Salesmen problems) can be efficiently transformed into K-bounded pseudo-Boolean functions.

Whitley notes that this class of functions has very useful properties for evolutionary optimization.  It has been proved that for single-bit mutations, the locations of bit-flips that will improve fitness can be computed in constant time.  Thus, from a given bit string, it is possible to very efficiently ``hill-climb'' to a local optimum.  Given this property, Whitley describes a highly efficient, \textit{deterministic} recombination method, called ``partition crossover'', which deterministically ``tunnels'' between different local optima in the fitness landscape.  The function can then be efficiently optimized via the combination of partition crossover and dynamic programming.

This combination of methods has been successfully applied to large combinatorial optimization problems such as scheduling and routing. These evolutionary computation methods seem very different from mechanisms in genetics and natural evolution.  However, Whitley notes that it is an open question whether biological evolution has perhaps found and exploited this optimization trick, and speculates about some possible connections, asking, for example: What if the fitness landscape of natural DNA is ``K-bounded''?  This is similar to a proposal made earlier by Stuart Kaufmann about his so-called NK-landscapes, which are inspired by genetic regulation in nature (Kauffman \& Weinberger, 1989). 

\noindent\textbf{Discussion:} 

Discussion centered around the applicability of the method to more general problems in evolutionary computation and the possibility of existing biological analogs. Some aspects of Whitley's algorithm seemed more applicable or more biologically plausible to the attendees. For example, since the algorithm involves recombination between individuals of similar fitness, one might argue that assortative mating (i.e., a preference for mating with individuals similar to oneself) might serve a similar function. While assortative mating might be a good trick in certain contexts, the analogy neglects another important component of the approach---its gray box optimization algorithm for assessing fitness, which requires some information about how the function we are optimizing maps input strings to output values. It was less clear how this information might be accessible to the mechanisms of biological evolution. Nevertheless, one commenter suggested that a very small amount of information could \textit{potentially} be leveraged to significantly improve evolution, so further investigation would be warranted even if this particular algorithm was biologically implausible. Finally, as commenters noted in Mike Lynch's talk, the designers of evolutionary algorithms have greater access to and control over the fitness function, so this kind of gray box optimization might be feasible in the context of evolutionary computation. 

\subsection[Discussant Comments (Michael Hochberg)]{Discussant Comments \& General Discussion:}

\begin{adjustwidth}{1cm}{1cm}
\onehalfspacing
\textit{Michael Hochberg is distinguished Research Director at the Centre National de la Recherche Scientifique at the University of Montpellier. Michael's research focuses on the ecology and evolution of disease, especially the within-host evolution of micro-parasites and cancer cells.}
\end{adjustwidth}	

Michael Hochberg began by noting that computational and biological approaches to evolution are akin to parallel universes---similar, but largely non-interacting. That said, he argued that this fact about contemporary scientific disciplines had more to do with the vagaries of history and less to do with the limits of evolutionary theory. Biological evolution is really just one instance of evolution, but biology become the face of evolution because it can lay claim to many of the earliest and greatest evolutionary theorists. This suggests that we can be optimistic about the fruitfulness of non-biological approaches to evolution, since evolutionary theory is bigger than any single scientific discipline. For example, Hochberg argues that evolutionary computation, unlike evolutionary biology, has the ability to cleanly isolate evolutionary processes for study.

Turning to the talks of the day, Hochberg suggested one overarching theme and three topics for further consideration. The overarching theme tying the talks together was the importance of scaffolding in evolution---or how existing structures (e.g. physical traits, biological processes, or elements of the environment) can facilitate the evolution of novel structures. This theme was especially clear in Bongard's talk, where he suggested that evolution capitalizes on naturally-occurring order. Subsequent discussion also addressed this theme, and in particular, how virtual environments might exhibit the rich space of possibilities afforded by natural environments. Commenters were especially interested in generating and identifying potentially useful adaptations in the vicinity of the organism's current phenotype. 

The first topic raised by Hochberg was whether evolutionary computation could experiment with the time scale of selection in ways not possible in biology. We normally think of evolution as a greedy or short-sighted process of optimization, selecting only on the basis of present phenotypes. It is less clear that evolutionary algorithms should be limited in this way. In principle, the designers of these algorithms could include fitness measures that attempt to ``look ahead'' and see how a trait might be beneficial in the future. The second topic was how we might better understand how fitness landscapes shift as the environment changes or as organisms gain new traits. The final topic was whether the accumulation of neutral or mildly deleterious mutations (especially those invisible to selection for the reasons laid out by Mike Lynch) could form the basis of future adaptive evolution. For example, a trait might be mildly deleterious in one environment, but significantly beneficial in another. As such, it may be benefit a species (in the long term) to have a diversity of potentially useful mutations upon which selection can later act. 

\subsection[``World Models and Attention for Reinforcement Learning'' (David Ha)]{``World Models and Attention for Reinforcement Learning''}

\begin{adjustwidth}{1cm}{1cm}
\onehalfspacing
\textit{David Ha is a Research Scientist at Google Brain. He studied math and engineering at the University of Toronto. David's work focuses on topics like neural networks, neuroevolution, and reinforcement learning.}
\end{adjustwidth}

\noindent David Ha surveyed his recent research and discussed how evolutionary algorithms, neural networks, and reinforcement learning can be combined to solve difficult problems. Appealing to work on the nature of consciousness and perception by Stanislas Dehaene and others, Ha argues that modeling plays an important functional role in cognition. Good models capture patterns in data and explain how superficially different events are instances of a simpler underlying regularity. For example, a detective might realize that the puzzling and apparently unconnected activities of a suspect were all in furtherance of a single criminal scheme. Similarly, in learning to drive in a simulated environment, a reinforcement learning agent would (ideally) learn that crashes subsequent to contacting different objects are all instances of collisions and should be avoided. 

Given this, Ha contends that models act as a kind of information bottleneck between perception and action. By enforcing the simplicity of these models, the agent is forced to build models that ``select, amplify, and propagate" task-relevant information. Memorizing the data and storing a complex look-up table of behavioral rules will not do. Ha details several instances of this ``bottleneck'' principle, each of which force the agent to identify the patterns that really matter for solving a problem. In the simplest case, a fairly literal bottleneck might be used. In this case, a neural network is forced to encode a visual scene (e.g., a top down view of a race car and the nearby track) using a small \textit{latent vector} of values. A recurrent neural network then learns to predict the next latent vector on the basis of the previous one. A reinforcement learning agent is then supplied with the latent vector and the neural network's prediction, which it uses to guide its behavior. In this way, the agent is forced to distill relevant information about a scene and then learn the patterns in how that information changes over time. 

Other bottlenecks are possible. For example, we might limit the ability of the agent to attend to various elements of the visual scene. To do this, we might allow a neural network to view the scene and decide which patches of the scene are most worth attending to. This \textit{attention} mechanism determines what patches are used for further processing. In a visual demonstration of the technique, Ha showed that agents learning to drive will (after training) selectively attend to salient details in the scene (e.g., walls in the path of the vehicle). A final method for forcing an agent to be more choosy involves distorting or scrambling the scene presented to the agent. This means that the agent cannot easily exploit superficial features of the scene but must carefully identify, reconstruct, and encode its most relevant features. The connection to evolutionary computation lies in the fact that everything from network architectures to agent policies can be designed through a process of variation and selection. In many cases, the problems that agents must solve do not allow us to train those agents via gradient descent methods. As such, evolutionary computation is an important tool for augmenting reinforcement learning and neural networks.

\noindent\textbf{Discussion:} 

The discussion of David Ha's talk touched on the importance of good data in reinforcement learning, the ability of his models to generalize to new cases, and the connection between his methods and evolutionary computation. The first commenter noted that giving agents information about random or arbitrary situations often leaves out examples of important situations and asked how Ha might generate better data. Ha responded that once an agent begins learning from random data, it develops a behavioral policy that leads it to encounter novel and (because they result from its own policy) more relevant situations. Of course, the agent learns from these novel situations, causing it to update its policy, to encounter more instructive situations, to update its policy again, and so on. One might also, Ha suggested, add a curiosity mechanism that encourages the agent to explore different situations from which it might learn. 

The second commenter worried that reinforcement learning agents might be over-fitting the training data, causing them to fail when (seemingly) small details about their environment are changed. Ha noted that there are at least two approaches one might take in such a situation. First, one might try to force the existing learning algorithm to learn more general lessons by augmenting data with a number of different changes (e.g., adding noise, scrambling blocks, changing textures, etc.). Second, one might change to the underlying algorithm. For example, one might use an algorithm which restricts an agent to learning structured representations of its data (e.g., hierarchically structured representations). This move could force the agent to develop useful abstractions of its data rather than latch on to details that will not generalize.  

Commenters also expressed interest in how Ha's reinforcement learning algorithms have/could be integrated with evolutionary approaches. Ha noted three ways of doing this. First, he noted that some attention mechanisms (e.g., for attending to some image patches and not others) are discrete, either attending or not attending to each thing. This discreteness challenges the gradient descent methods used to train, say, neural networks which learn by making many finely graded adjustments to the network's weights. Evolutionary algorithms are not limited in this way. Second, Ha noted that the rise of cheap CPUs with many cores had lowered the technological barrier to wider application of evolutionary methods. Finally, Ha argued that the evolutionary concepts other than selection might be useful. For example, genes \textit{indirectly} encode traits by specifying a recipe for their construction. This seems to constrain evolution in helpful ways, and it might be possible to indirectly represent and evolve important elements of reinforcement learning agents.  

\subsection[``Evolution and Evolutionary Computation'' (David Krakauer)]{``Evolution and Evolutionary Computation: A Few Puzzling Questions''}

\begin{adjustwidth}{1cm}{1cm}
\onehalfspacing
\textit{David Krakauer is President and William H. Miller Professor of Complex Systems at the Santa Fe Institute. His research concerns (among other things) the evolution of intelligence, collective intelligence, and the relationship between intelligence and the laws of biology and physics.}
\end{adjustwidth}

\noindent David Krakauer outlined four broad questions for evolutionary computation intended to clarify the goals and nature of the field. The first question concerns the phylogeny of adaptive processes. Krakauer notes a deep mathematical similarity between evolution by natural selection and a number of other processes that we might think of as adaptive. Examples include the mathematics of how game theoretic equilibria are reached or how our probability assignments evolve as more observations are made. Curiously, these kinds of processes are highly relevant to how evolved living systems interact with their environments and other living systems. A key question then is: What relationship (if any) is there between the evolutionary processes that gave rise to living systems and these closely related processes? 

The second question is whether the literature on evolutionary computation has overemphasized the role of natural selection as an optimization process, neglecting a richer family of related concepts. In particular, Krakauer has in mind concepts such as evolvability, robustness, and resilience---each of which is an important feature of evolved strategies. A further question is how these concepts fit in with the idea of natural selection as a computational process. If the computational work of evolution is function optimization, then how should we understand these other consequences of evolution? Perhaps we should expand our sense of what computation evolution is performing or introduce novel non-computational concepts to further our understanding of the evolutionary process. 

The third question concerns the kinds of computers that arise from evolutionary processes. In particular, does evolution favor any particular kind of computer over others? For example, some computers are quite general purpose while others are more specialized. The most familiar computers built by evolution are the neural networks implemented in animal brains. That said, computational processes can also be seen at the molecular, genetic, and social levels. As Krakauer argues, the physical realizations of these processes are very different in scale and composition, yet they share a surprising number of characteristics. For example, they involve collectives of stochastic components, they operate asynchronously without centralization, they are assembled as needed, and they are robust to perturbations. This raises the question of whether this kind of collective computation is particularly favored by evolution or whether it is simply better for many jobs than the kind of computation that has been engineered in electronic computers.

The final question is whether evolutionary computation is primarily concerned with evolution as a computational process or in understanding living systems more generally. We might think of this in terms of the distinction between work on evolutionary algorithms and work on artificial life. The latter is more concerned with how the distinctive characteristics of life arise from evolution, development, and other biological process that we are able to computationally model. The former is more concerned with using evolution to solve particular problems. As Krakauer argues, there are likely trade-offs involved in choosing which sort of research project we will pursue. Artificial life, for example, is less likely to find solutions to the exact problems that interest us, but the solutions it discovers might be more diverse or more robust given its richer and more realistic approach. Again, this connects with a key theme of the workshop---the promise of harnessing the distinctive advantages of evolution as it appears in nature. 

\noindent\textbf{Discussion:} 

The discussion, much like the talk, covered a broad range of relevant and interesting topics. One commenter asked what the signatures of different optimization processes might be. Krakauer argued that this turns on important open questions about how best to coarse-grain the target systems. Another commenter wondered whether the distinction between computers of different generality (as framed in the talk) captured a relevant difference since simple modifications to specialized computers can give them the formal properties necessary for generality. Krakauer acknowledged that that this kind of sharp distinction was inappropriate, but also argued that we need better ways of describing the continuum from the simplest calculator to highly general computers. 

One commenter responded to the idea that evolutionary computation has focused too much on optimization. The commenter argued that this reflected the field's origin in engineering. However, they also noted how more recent work has argued that many ways of improving evolutionary algorithms with respect to optimization have moved them further from their biological roots and sacrificed other interesting properties of biological evolution. Next, the discussion turned towards the kinds of design choices that biological evolution tends to make. For example, Krakauer argued that evolution often tries to offload work from development and learning and build solutions directly into the organism. Naturally, being born with specialized tricks for surviving and reproducing reduces the burden on development and learning. More generally, evolution seems to favor simpler and more specialized mechanisms wherever possible.

\subsection[``Artificial Adversarial Intelligence'' (Una-May O'Reilly)]{``Artificial Adversarial Intelligence''}

\begin{adjustwidth}{1cm}{1cm}
\onehalfspacing
\textit{Una-May O'Reilly is the Founder and Principal Research Scientist of the AnyScale Learning for All (ALFA) Group at MIT-CSAIL. Her research applies machine learning and evolutionary algorithms to knowledge mining, analytics, and optimization.}
\end{adjustwidth}

\noindent Una-May O'Reilly discussed her work on adversarial learning, highlighting the common failure modes of evolutionary and non-evolutionary adversarial methods. In adversarial learning, two or more agents with incompatible goals explore a rich space of possible strategies for achieving those goals. A successful strategy is one that works well for an agent given the strategies employed by its opponents. Of course, this competitive context is ever-changing since the agent's opponents are also adapting their strategies. This competitive ``arms race'' will \textit{ideally} lead to better and better strategies---just as methods of counterfeiting and counterfeit detection will improve over time as each side attempts to stay ahead of the competition. 

In practice, this kind of mutual improvement is not always seen. For example, if an agent's present strategy works well against its opponent's present strategy but not against its opponent's \textit{previous} strategy, its opponent can get ahead by reverting to its previous strategy. More generally, if agents are not able to remember and are no longer exposed to older strategies, agents may \textit{cycle} through strategies instead of developing better and better strategies. Similar failure modes can occur when an agent's strategy becomes to narrowly tailored to combat its opponent's strategy. This can lead to brittle strategies that fail when paired with new opponents. Finally, it is possible for an agent to so thoroughly dominate the competition that none of the variations on its opponents strategies do better than any others. This shuts down the process of developing better strategies through gradual adaptation. 

These failure modes can arise whether or not the agents learn over successive generations of an evolutionary algorithm or over successive steps of gradient descent. A key component in avoiding these failure modes, O'Reilly argues, is to encourage diversity. Each of the failure modes results in the agent's being locked into a particular strategy (or cycle of strategies). If opponents could be encouraged to explore more widely, these failures might be mitigated. For example, if an agent's opponents employ multiple strategies (e.g., some old and some new), the agent cannot succeed by learning a strategy that only succeeds against a single strategy (e.g., the newest and best). 

O'Reilly details one method for encouraging this kind of strategy diversity---adding spatial structure to a competitive environment. When every agent competes against every agent, the single best strategy (at any given time) will be favored. However, if agents compete against their neighbors only, then multiple locally successful strategies can arise. These strategies can still encounter more distant strategies, but only as quickly as they can spread through the environment (i.e., to neighbors of neighbors of neighbors, etc.). This latency allows greater diversity since new strategies can arise and even be refined before encountering strategies refined elsewhere. Further, for any strategy to become dominant, it must be successful against all the local variations on alternative strategies.  

O'Reilly and her collaborators applied this spatial method to generative adversarial networks for image generation. These networks pit image generators against image discriminators that try to separate generated images from real-world images. These networks are notoriously difficult to train and frequently exhibit the failure modes described above. Normally, one generator competes against one discriminator, but O'Reilly created a spatially-distributed population of generators and discriminators and found that this substantially reduced the failure modes endemic to adversarial learning. These results further reinforce the importance of encouraging algorithms to explore a diverse range of possible solutions---a major theme of the workshop overall. 

\noindent\textbf{Discussion:} 

The discussion focused primarily on finding ways to use O'Reilly's methods to address other challenges in machine learning. For example, one commenter noted that adding a spatial component to an evolutionary algorithm increased the interpretability of the solutions it discovered. O'Reilly replied that she was not yet sure how to represent her models in a way that would benefit from this effect, but added that the spatial component had made some aspects of the evolutionary process more interpretable (e.g., how fitness propagates in a population). Another commenter noted that generative adversarial networks are know for their creativity (i.e., their ability to produce realistic images that are not mere copies of their training data). The commenter then suggested that spatial evolution might foster greater creativity, which would represent another advantage of her approach. 

\subsection[Discussant Comments (Aleksandra Faust)]{Discussant Comments \& General Discussion:}

\begin{adjustwidth}{1cm}{1cm}
\onehalfspacing
\textit{Aleksandra Faust is a Staff Research Scientist and Reinforcement Learning research team co-founder at Google Brain Research. Her research interests include safe and scalable reinforcement learning, learning to learn, motion planning, decision-making, and robot behavior.}
\end{adjustwidth}	

Faust began the discussion with a brief synopsis of the talks before proposing several questions for general discussion. One question was whether work like David Ha's suggests a way of evaluating different cognitive architectures with respect to the degree of abstraction they involve (e.g., the degree of abstraction in an agent's latent space representation of the problem domain). Another questions was whether (and why) spatial approaches to evolution might help to prevent populations from getting stuck in local minima. Finally, she reiterated David Krakauer's questions about the nature and aims of evolutionary computation (e.g., What is the object of study? What does evolution do if not optimization?). 

The general discussion focused primarily on this last question. One issue which several attendees weighed in on was how to describe the way that evolution improves systems without appealing to optimization. The idea of \textit{satisficing} (i.e., settling for good enough solutions) is a natural alternative to optimization (i.e., trying to find the best solution). For example, one commenter suggested that adversarial contexts make this characterization particularly natural because a trait or strategy will be favored so long as it is better than those employed by adversaries. Of course, if relative fitness (rather than victory) determines whether and how much one gets to reproduce, then winning strategies may not be equal. As other commenters suggested, engineers using evolutionary algorithms often seek designs that work better than the state of the art instead of seeking optimal designs. Hence, the aim here is continual improvement by a set of steps none of which attempt to fully optimize the design. Further complicating matters, engineers often use evolved designs as a starting point for human design.

Another commenter argued that while evolution does engage in some optimizations, there are many optimization algorithms that lack striking properties of evolution (e.g., creativity).  It might be better, then, to focus on understanding those aspects of evolution that set it apart from other methods rather than deciding how best to characterize the way it improves the systems upon which it acts. This suggestion raised the topic (discussed repeatedly throughout the workshop) of what conditions are necessary for evolution to exhibit its distinctive attributes. Faust made a particularly interesting proposal about one such condition. She noted that the environment does not merely change over time---it is changed by organisms themselves and these changes shape the fitness landscape faced by their descendents. As such, she suggests incorporating more ways in which simulated organisms can change their environment in persistent ways. Along similar lines, one commenter noted that evolutionary arms races require changing cost functions in order to continue driving change in the competing species. In both cases, the idea is that the creative diversity of evolution can be driven by dynamic elements of the context in which the evolving species are embedded. 

\section{Conclusion}
The workshop's participants offered a breadth of interesting, relevant, and informed ideas about evolution and evolutionary computation. These ideas can be grouped under two broad themes. First, a number of the ideas concerned the nature and aims of the field of evolutionary computation. Second, a number concerned the nature of evolution as a process and the differences between biological evolution and evolutionary computation. In this conclusion, we will summarize the most important of these ideas and discuss future directions for the field. 

\subsection{The Nature and Aims of Evolutionary Computation}

During the workshop, there was a great deal of reflection on the field of evolutionary computation, spurred by observations from those in the field and by questions from their interdisciplinary colleagues. One concern was about the centrality of \textit{computation} to the field. Rod Brooks, for example, suggested that our idiosyncratic models of computation may blind us to other ways of thinking about intelligence. Josh Bongard echoed these thoughts, arguing that evolution often relies on exploiting naturally-occurring order, offloading some of the hard work of life and intelligence to the body and environment. Approaching the question of computation somewhat differently, David Krakauer sought clarification about the difference between evolutionary computation and artificial life, where the former seems to concern evolution as a computational process and the latter seems to concern life as revealed via simulated organisms. In other words, Krakauer asked if computation is primarily a subject of study or a tool for studying something else. 

Another important issue was the relationship between evolutionary algorithms and other machine learning methods, such as deep learning and reinforcement learning. Presenters touched on all these areas, and there were often questions about where evolutionary approaches belong in the larger toolbox of machine learning. As mentioned above, one distinctive advantage of evolutionary algorithms is that they do not rely on gradient descent and (hence) on differentiable loss functions. Another reason one might choose an evolutionary algorithm concerns the kinds of solutions these algorithms learn. For example, several speakers noted that evolution is more conducive to generating a diverse set of solutions to problems. A major question, however, was the extent to which these features can be incorporated into other methods (e.g., by a loss function that rewards diversity). Overall, the speakers and participants seemed to prioritize solving problems and understanding intelligence over championing any particular method. That said, there was a decided optimism about the continuing relevance of evolution to machine learning, since (if nothing else) it has produced forms of intelligence far more advanced than anything our current methods allow. 

In a similar vein, participants were also interested in how the broader methodology of evolutionary computation compares to the methodology of other areas of machine learning. More than one participant observed that new work in evolutionary computation rarely builds on the products of earlier work. This isn't to say that earlier \textit{ideas} do not inform later ones, but rather that the specific solutions learned in one context are rarely used or built-upon to solve other problems. This kind of transfer between contexts is more common in deep learning where new algorithms often start with features learned previously (see e.g., the extensive use of embeddings from large pre-trained language models). One explanation for these differences is that evolutionary computation has, as yet, few standardized ways of representing solutions, and this inhibits re-using prior solutions or parts of prior solutions. In deep learning, the generality of neural networks typically makes it easy to incorporate pre-trained networks into new models so long as they deal with roughly the same type of data. Another explanation is that, as a field, evolutionary computation is more interested in understanding how evolution works than in using it to solve practical problems. On this view, much of the work in evolutionary computation is basic research. While the general consensus seemed to be that methods for building on prior work were desirable, there was still a sense that important aspects of the evolutionary process remain to be discovered through basic research. 

\subsection{The Distinctive Properties of Evolution}

The second major theme concerned the distinctive properties of evolution as a design or optimization process. On balance, workshop attendees seemed to view biological evolution as having many distinctive advantages---advantages which are less conspicuous in existing evolutionary algorithms. For example, an important advantage of biological evolution is that it designs organisms with the help of an extremely rich and varied environment. Such environments subject agents to a more diverse set of possible circumstances, leading to more generalizable solutions. They can also change indefinitely in response to changes in the agents. The ensuing feedback loop between agent and environment seems to be a major driver of novelty and diversity in evolution. Simulated environments, in contrast, are still too limiting to realize the full potential of evolution. For example, Ken Stanley noted that while his maze evolution algorithm generated increasingly complex and diverse mazes, it would never give rise to anything other than mazes. In contrast, biological organism often dramatically reshape their environments upon acquiring new traits (see e.g., the oxygenation of Earth's atmosphere after the evolution of photosynthesis). Another distinctive advantage of biological evolution is its immense scale relative to evolutionary algorithms. It is difficult to fathom the vast scale of, say, microbial evolution in the Earth's oceans. Even the evolution of larger organisms with smaller population sizes involved more organisms and more generations than could be simulated on human time scales. 

Despite these advantages of biological evolution, there are several advantages to simulated evolution. For example, it is far easier to isolate particular evolutionary processes for study. The complexity of natural environments makes it difficult to, say, identify important causal factors or explain the adaptive advantage of a particular phenotype. Also, simulated evolution need not exhibit the myopia of biological evolution. Simulated fitness functions can select organisms to reproduce on the basis of any arbitrary characteristic (e.g., the value of its traits in the far future or the value of its traits to the species as a whole). Hence, simulated evolution has non-trivial advantages as both an investigative tool and as a design process.

More generally, participants were interested in the distinctive advantages of evolution over other methods of learning or design. Ken Stanley spoke at length about the open-endedness of evolution (i.e., its tendency to produce endless diversity and complexity). Other processes, he argued, were more apt to produce convergence and stagnation. David Krakauer argued that viewing evolution as an optimization algorithm alone failed to do it justice. For example, evolution appears to discover robust and resilient solutions---features that might be neglected by simply minimizing some fixed loss function. This robustness likely has more to do with constantly shifting fitness landscapes than with the effectiveness of evolution as a method of hill-climbing. Other advantages of evolution are not quite as distinctive, but are particularly pronounced in evolution. For example, as Una-May O'Reilly showed, adversarial feedback loops often drive evolution. Further, as Josh Bongard and Ken Stanley argued, environmental feedback loops are also a major driver of evolutionary change, and even without change in the environment, the space of possibilities open to an organism is constantly changing as it acquires and loses traits. More broadly, we might say that evolution involves dynamic individuals, populations, and environments, and this dynamism facilitates the kind of feedback loops that drive evolution. Evolution also  elegantly incorporates constraints from the environment (e.g., finite resources), the body (e.g., developmental mechanisms), and other organisms (e.g., competitors). These constraints narrow an unwieldy space of design possibilities, guiding evolution toward robust solutions that take advantage natural order and succeed in the presence of resource limitations and adversaries. 

\subsection{Frontiers in Evolutionary Computation}

With these themes in mind, we can turn our attention to promising directions for future research in evolutionary computation. These research frontiers all focus on realizing the obvious, yet largely nascent, potential of evolutionary algorithms. The most straightforward path to realizing this potential involves \textit{scaling up} existing approaches. Better hardware and software has massively advanced the field of deep learning by facilitating larger models and faster training. Similar technological advances for evolutionary computation could substantially improve results without requiring substantial changes to existing algorithms. Fortunately, speakers seemed optimistic about the falling costs of CPUs with many cores, the availability of highly parallel super computers, and the (possible) development of bespoke hardware for evolutionary computation. Such advancements would allow researchers to scale up evolutionary computation in the years to come. Besides realizing the potential of existing algorithms, more computational power would also aid research along other frontiers by reducing iteration time and costs and, crucially, by allowing simulations with richer structure (e.g., more complicated topologies) and more realistic environments (e.g., with the capacity to change significantly over time). That said, some attendees urged caution about viewing greater scale as a panacea and encouraged everyone to think more deeply about other processes central to evolution (e.g., development). 

Another important frontier for research is blending evolutionary approaches with other machine learning paradigms. Many familiar technologies gained a foothold in science by way of their practical applications. Rocketry, for example, was not seriously funded by the US government until it was marketed for what now seems like a minor auxiliary function---helping planes take off from short runways (Clark, 1972). While evolutionary computation is more mature than rocketry was at this stage, finding ways to incorporate evolutionary algorithms into other algorithms will only increase interest in and funding for research in evolutionary computation. Besides these pragmatic considerations, it seems evident that nature, too, has blended evolution with other methods. For example, over short time scales, animals learn via processes akin to reinforcement learning. Over longer time scales, animals learn via natural selection. The latter provide organisms with a set of capacities and behaviors adapted to slowly varying elements of the organisms' environment. The former allow organism to adapt to elements of their environment that change too quickly for selection to act upon them. In David Ha's presentation, for example, he noted that evolution could be used to develop important components of models that improve reinforcement learning (e.g., discrete attention mechanisms).  

A third frontier for research involves enhancing the richness of simulated agents and environments. This might include, as one commenter suggested, simulating environments at finer grains to allow more diverse and realistic emergent phenomena. It might also include a richer simulation of the agent's body or between-agent interactions. This frontier emphasizes the fact that, in the real world, agents are both embodied and situated in an environment---a recurring idea in the workshop. As discussed earlier, these factors constrain evolution in useful ways, and richer environments promote open-ended evolution and involve emergent order that agents can use to their advantage. In addition to adding more details to simulated agents and environments, it seems important to identify the most important details and fine tune them to optimize evolutionary algorithms. For example, Mike Lynch discussed the importance of population size, random drift, and recombination rates to evolution. Each of these things are under the control of researchers designing evolutionary algorithm, and getting the right values could markedly improve the solutions discovered by our algorithms. 

Overall, the workshop sparked lively and enlightening discussions about the nature and future of evolutionary computation. It also brought together biologists and computer scientists to share their insights about evolution as they study it. By reflecting on the field and sharing their research, the attendees furthered a deeper understanding of evolution and generated promising ideas about how to advance evolutionary computation. Despite the challenges of scaling up and refining evolutionary computation, the attendees shared a sincere excitement about the field and its role in the future of artificial intelligence. 

\subsection*{Acknowledgments}
The workshop was funded by a grant from the National Science Foundation (\#2020103) as part of the Foundations of Intelligence in Natural and Artificial Systems project at the Santa Fe Institute.

\newpage

\section{References}

\begin{flushleft}

\begin{list}{}
{\leftmargin=1em \itemindent=-1em \itemsep=-.4em}

\item Clark, J.D. (1972). \textit{Ignition!: An Informal History of Liquid Rocket Propellants}. New Brunswick, NJ: Rutgers University Press. 

\item Darwin, C. (1872). \textit{The Origin of Species}. 6th Ed. London: John Murray.  

\item Kauffman, S.A. \& Weinberger, E.D. (1989). ``The NK Model of Rugged Fitness Landscapes and its Application to Maturation of the Immune Response.'' \textit{Journal of Theoretical Biology}. 141(2). 211-245.

\item Miikulainen, R. \& Forrest, S. (2021). ``A Biological Perspective on Evolutionary Computation.'' \textit{Nature Machine Intelligence}. 3. 9-15.

\end{list}
\end{flushleft}

\end{document}